\documentclass[letterpaper, 10pt, conference]{ieeeconf}  % 美国信纸格式

\IEEEoverridecommandlockouts    % 允许使用\thanks命令
\pdfoutput=1

\usepackage{graphics}           % 处理PDF等位图
\usepackage{epsfig}             % 处理PostScript图形
\usepackage{times}              % 使用Times字体
\usepackage{amsmath}            % 数学公式支持
\usepackage{amssymb}            % 数学符号支持
\usepackage{balance}            % 平衡双栏内容
\usepackage{cite}               % 引用处理
\usepackage{graphicx}           % 图形处理
\usepackage{physics}            % 物理公式宏包
\usepackage{xcolor}             % 颜色支持
\usepackage{tikz}               % 矢量图形绘制

\usepackage{hyperref}           % 超链接支持
\usepackage{caption}            % 图表标题定制
\usepackage{subcaption}         % 子图支持
\usepackage{lipsum}             % 生成示例文本（实际使用时删除）
\usepackage{booktabs}           % 美观表格线
\usepackage{tabularx}           % 自适应宽度表格
\usepackage{wrapfig}            % 图文混排
\usepackage{multirow}           % 表格多行合并
\usepackage[section]{placeins}  % 控制浮动体位置
\usepackage{siunitx}            % 角度符号

\title{\LARGE \bf
OmniDP: Beyond-FOV Large-Workspace Humanoid Manipulation with Omnidirectional 3D Perception
}

\author{
Pei Qu$^{1*}$, Zheng Li$^{1*}$, Yufei Jia$^{2}$, Ziyun Liu$^{1}$, Liang Zhu$^{1}$,
Haoang Li$^{1}$, Jinni Zhou$^{1}$, Jun Ma$^{1,3\dag}$%
\thanks{*Equal contribution, \dag Corresponding Author.}
\thanks{$^{1}$The Hong Kong University of Science and Technology (Guangzhou).
        {\tt\small \{pqu458, zli514, zliu176, lzhu686\}@connect.hkust-gz.edu.cn, 
        \{haoangli, eejinni, eejma\}@hkust-gz.edu.cn}}%
\thanks{$^{2}$Tsinghua University.
        {\tt\small jyf23@mails.tsinghua.edu.cn, zhouguyue@air.tsinghua.edu.cn}}
\thanks{$^{3}$The Hong Kong University of Science and Technology.
        {\tt\small jun.ma@ust.hk}}
}
\begin{document}
% 重定义双栏环境以包含 teaser 图
\maketitle
% 正文部分

\begin{abstract}

The deployment of humanoid robots for dexterous manipulation in unstructured environments remains challenging due to perceptual limitations that constrain the effective workspace. In scenarios where physical constraints prevent the robot from repositioning itself, maintaining omnidirectional awareness becomes far more critical than color or semantic information.
While recent advances in visuomotor policy learning have improved manipulation capabilities, conventional RGB-D solutions suffer from narrow fields of view (FOV) and self-occlusion, requiring frequent base movements that introduce motion uncertainty and safety risks. Existing approaches to expanding perception, including active vision systems and third-view cameras, introduce mechanical complexity, calibration dependencies, and latency that hinder reliable real-time performance. 
In this work, We propose OmniDP, an end-to-end LiDAR-driven 3D visuomotor policy that enables robust manipulation in large workspaces. Our method processes panoramic point clouds through a Time-Aware Attention Pooling mechanism, efficiently encoding sparse 3D data while capturing temporal dependencies. This 360° perception allows the robot to interact with objects across wide areas without frequent repositioning. 
To support policy learning, we develop a whole-body teleoperation system for efficient data collection on full-body coordination. Extensive experiments in simulation and real-world environments show that OmniDP achieves robust performance in large-workspace and cluttered scenarios, outperforming baselines that rely on egocentric depth cameras.

\end{abstract}

\section{Introduction} \label{sec 1}

Humanoid robots capable of dexterous manipulation in unstructured environments have long been a central goal in robotics. Recent advances in visuomotor policy learning, particularly imitation learning~\cite{2024aloha,2024dreamitate,2025visualImitation} and diffusion-based action generation~\cite{2025dp,2024dp3,2025idp3}, have substantially enhanced humanoids’ ability to learn manipulation skills directly from visual observations. By learning end-to-end sensorimotor mappings, these policies bypass explicit state estimation and task-specific modeling, enabling flexible skill acquisition across embodiments~\cite{2024octo}. This progress opens opportunities for deploying humanoid robots in service robotics~\cite{2024humanoidAssistance}, household assistance~\cite{2025behaviorRobotSuite}, and collaborative human–robot interaction~\cite{2025cola}.

    \begin{figure}[t]
        \centering
        \vspace{0.2cm}
        \includegraphics[width=0.48\textwidth]{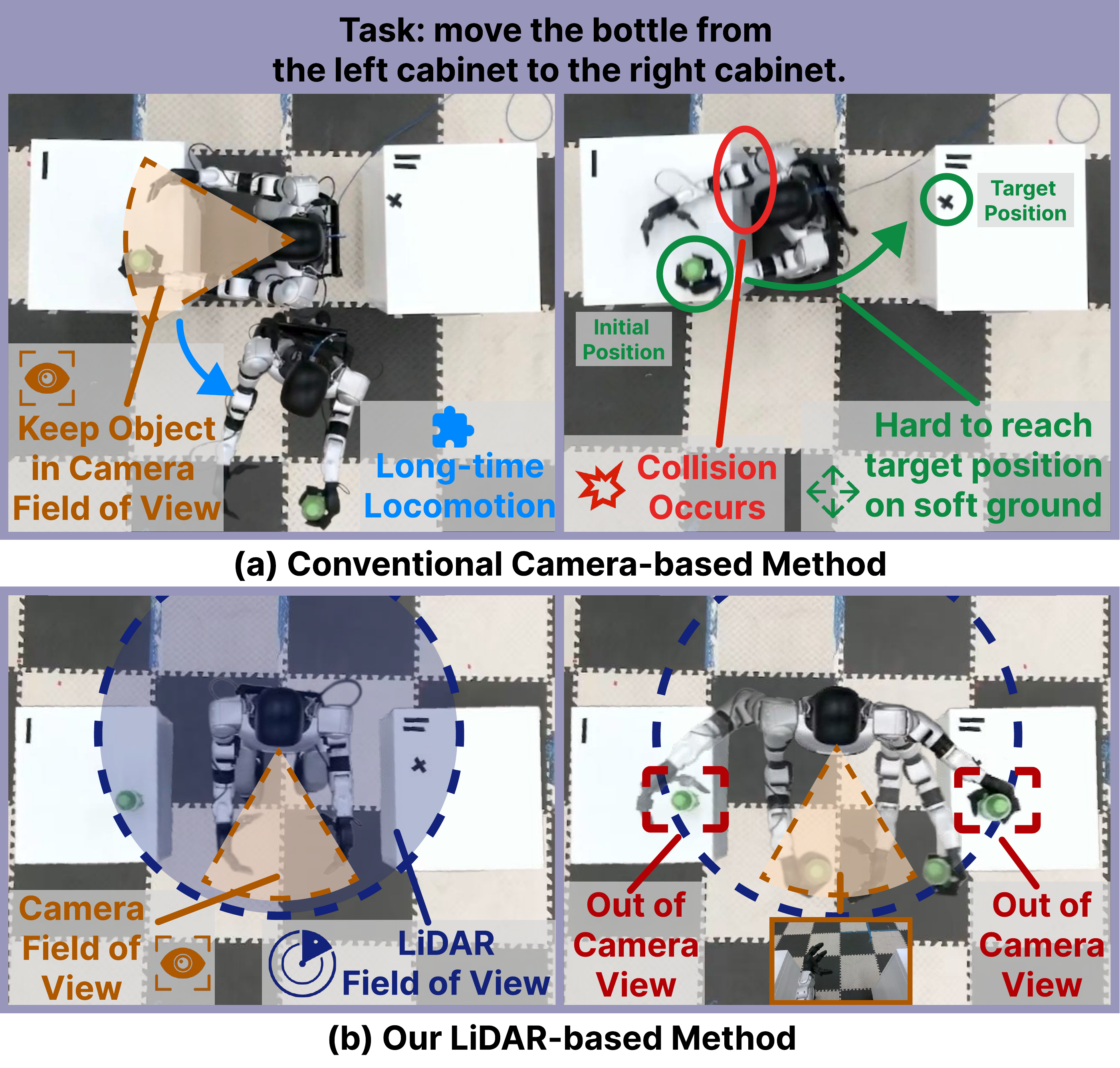}
        \caption{\textbf{OmniDP.} (a) The narrow field of view of RGB-D cameras prevents perception of objects outside visual range, causing task failures and collisions in space-constrained environments where robot repositioning is difficult; (b) by utilizing panoramic 3D LiDAR perception, our approach empowers humanoid robots with the capability to perform manipulation tasks across a large workspace, including areas outside the camera's visual blind spots.}
        \label{fig:teaser}
    \end{figure}

Despite recent advances in visuomotor policies, perceptual limitations persist in humanoid robots operating in unstructured environments. 
% In such environments, expanding the effective manipulation workspace is critical, as target objects are often distributed across a large 3D space~\cite{2024aloha,2025robopanoptes}. Limited perception forces frequent base movement to compensate for blind spots, introducing motion uncertainty and safety risks. 
In scenarios where the robot's base is restricted in movement and rotation due to physical constraints, target objects may remain undetected by conventional RGB-D cameras, thereby limiting the effective workspace to what is immediately visible~\cite{2024aloha,2025robopanoptes}. In these "viewpoint-limited" cases imposed by standard visual sensors, omnidirectional awareness becomes more critical than color or semantic information.
Mainstream RGB-D solutions rely on single-view input, which suffers from a narrow field of 
view (FOV) and self-occlusion~\cite{2023robotStructure,tian2024viewinvariant,2024robotSeeRobotDo}. Moreover, the commonly used depth-to-point-cloud approach, while providing 3D geometry, remains inherently viewpoint-constrained~\cite{2024pcdImprove}. Even introducing third-view cameras fails to resolve this, as they depend on extrinsic calibration and struggle with camera pose generalization~\cite{2023visualPolicy,2025PLKCalib}. These limitations hinder the reliable deployment of humanoid robots for large-workspace, continuous operations in unstructured environments.

% Define "field of view limitation" and explain the need of using LiDAR
% In scenarios where the robot's base is restricted in movement and rotation, target objects may remain undetected by the current visual perception, even within the robot's manipulation range. In these "viewpoint-limited" cases, omnidirectional awareness becomes more critical than color or semantic information. The ability to perceive 360° of the surrounding space without frequent repositioning is essential for efficient task execution, as spatial coverage and obstacle detection take precedence over object color or semantics.
% LiDAR is an ideal choice for such environments, as it provides dense 3D point cloud data with a wide and consistent field of view, overcoming the viewpoint limitations of RGB-D cameras.

% In such scenarios where the robot's base is restricted in movement and rotation, target objects can remain undetected by the current visual perception even though they are within the robot's potential manipulation range. In these "viewpoint-limited" cases, omnidirectional awareness becomes far more critical than color or semantic information. The robot's ability to perceive the full 360° of its surrounding space without the need for frequent repositioning is essential for safe and efficient task execution, as relying on object color or semantics can be secondary to spatial coverage and obstacle detection.

Several approaches have been explored to expand the effective workspace for robotic manipulation. Active vision systems, which adjust viewpoint through camera motion, can in principle extend perception range~\cite{2025visionInAction, 2025egomi, chuang2025active}. However, for humanoid robots, this often requires additional degrees of freedom in the head or neck, introducing mechanical complexity, higher costs, and control overhead. Moreover, active movements may interfere with ongoing task execution and introduce latency, limiting real-time performance. Panoramic sensors, including $360^\circ$ cameras and LiDAR, provide wide horizontal fields of view without mechanical motion. Although they have shown promise in autonomous driving~\cite{2025dur360bev,2024fisheyebevseg}, navigation~\cite{2024omnidirectional,2025yuto,2024learning}, and locomotion~\cite{2025omniperception}, their use in end-to-end visuomotor manipulation for humanoid robots remains under-explored, representing a gap in enabling large-workspace manipulation tasks~\cite{2024learning,2024navila,2025safeLocomotion}.

To overcome these challenges, we propose a novel method for robust, large-workspace humanoid manipulation. First, we introduce \textbf{OmniDP}, an end-to-end LiDAR-driven 3D visuomotor policy that processes panoramic point clouds with a Time-Aware Attention Pooling mechanism, enabling efficient encoding of sparse 3D data while capturing temporal dependencies. This design expands the effective manipulation space through 360° perception, allowing the robot to perceive and interact with objects located outside the camera's FOV without frequent repositioning, as demonstrated in Fig.~\ref{fig:teaser}. Second, to support policy learning in this setting, we develop a whole-body teleoperation system tailored for humanoid robots, enabling efficient collection of large-workspace demonstration data that captures complex full-body coordination. Extensive experiments demonstrate that our method achieves robust performance in large-workspace scenarios requiring extended spatial coverage, as well as in cluttered environments where panoramic perception is essential for collision-free manipulation, consistently outperforming vision-based baselines.

Our core contributions are summarized as follows:
\begin{itemize}

\item We propose OmniDP, an end-to-end LiDAR-driven 3D visuomotor policy with time-aware encoding of panoramic point clouds, enabling beyond-FOV large-workspace humanoid manipulation.
\item  We develop a whole-body teleoperation system that enables efficient collection of human demonstration data involving coordinated whole-body movements.
\item We validate our method through extensive experiments in simulation and the real world, demonstrating robust large-workspace and omnidirectional collision-free manipulation in challenging environments.

\end{itemize}

\section{Related Work}\label{sec 2}

\subsection{Visuomotor Policy}
Recent progress in robot learning has been driven by visuomotor policies that map visual observations to actions, with approaches varying by representation type: 2D RGB, 2.5D RGB-D, and 3D point clouds.

\textbf{Approaches with 2D and 2.5D Visual Inputs.} Diffusion Policy~\cite{2025dp} pioneered the integration of diffusion models into robotic policy learning, generating action sequences from RGB observations to enable multimodal behavior synthesis in complex scenarios. Beyond imitation learning, vision-based reinforcement learning has also achieved significant progress in real-world settings\cite{2025RLmani,2025cotraining}. RGB-D observations, by incorporating depth information, provide geometric perception for contact-rich dexterous manipulation tasks. For instance, monocular metric depth alignment methods~\cite{2025monocular} can recover 3D scene structure without requiring dense depth annotations. However, constrained by fixed viewpoints and limited observation ranges, these methods are typically applicable only to local manipulation areas and struggle to develop a comprehensive global 3D scene understanding.

\textbf{Approaches with 3D Visual Inputs.} Approaches with 3D Visual Inputs. In contrast to 2D methods, point cloud representations explicitly encode 3D spatial geometry, offering richer structural information for complex manipulation tasks. 3D Diffusion Policy (DP3)~\cite{2024dp3} successfully extended diffusion models to point cloud representations, with subsequent research applying this framework to humanoid robotic manipulation ~\cite{2025idp3}. Regarding robustness, notable progress has been made through techniques such as implicit scene supervision~\cite{2025ISSpolicy} and variational regularization~\cite{2026informationFiltering}. Addressing the challenge of viewpoint sensitivity, ManiVID-3D~\cite{2025manivid} learns view-invariant representations via feature disentanglement, significantly enhancing model generalization across multiple viewpoints. Nevertheless, current mainstream methods still rely on point clouds acquired from fixed RGB-D cameras, inheriting limited fields of view and requiring accurate extrinsic calibration for reliable 3D reconstruction.

\subsection{LiDAR-based Perception}
LiDAR is an increasingly adopted exteroceptive modality for legged and humanoid robots due to its illumination invariance and reliable 3D geometry, supporting both reactive control and long-horizon autonomy.

Omni-Perception ~\cite{2025omniperception} learns omnidirectional collision avoidance for legged locomotion by consuming temporal LiDAR point clouds and directly producing control actions, reducing reliance on hand-crafted intermediate maps and emphasizing sim-to-real transfer with LiDAR-centric simulation support. 
LVI-Q ~\cite{2025lvi-q} tightly couples LiDAR with vision, inertial, and kinematic cues to obtain robust odometry on quadrupeds, mitigating drift under aggressive motion and partial observability, and providing a reliable state estimate for downstream navigation. 
BeamDojo ~\cite{2025beamdojo} targets agile humanoid locomotion on sparse footholds and leverages onboard LiDAR-based elevation mapping at deployment time, demonstrating the practical utility of structured terrain maps for precise stepping.
PILOT ~\cite{2026pilot} uses LiDAR as the sole external sensor input for environment perception. While LiDAR-based perception has demonstrated promising applications in locomotion and navigation, its potential in humanoid manipulation remains largely unexplored, particularly in tasks that demand large workspace operation and omnidirectional obstacle avoidance during arm deployment.

\section{Method}\label{sec:method}

\begin{figure*}[t] 
\centering
\vspace{0.3cm}
\includegraphics[width=\textwidth]{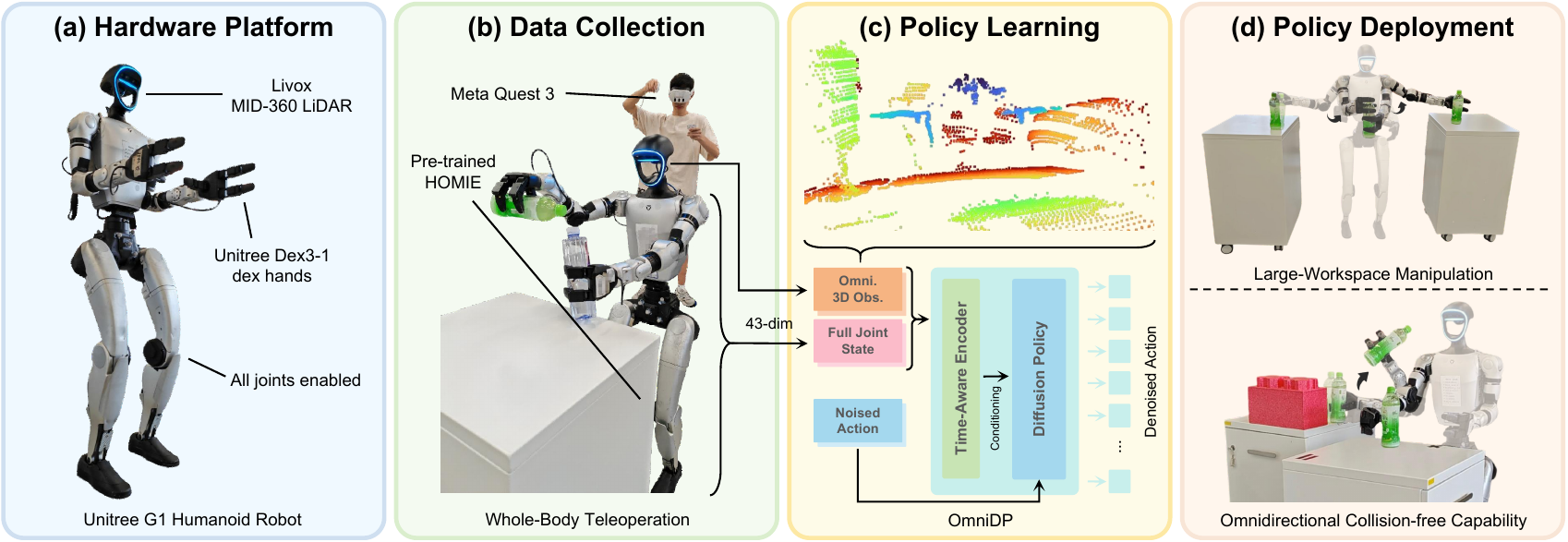} 
\caption{\textbf{Overview of system architecture.} Our system consists of four key components: (i) the panoramic perception hardware platform based on a Unitree G1 humanoid robot equipped with a Mid-360 LiDAR; (ii) a whole-body teleoperation system for efficient demonstration data collection; (iii) an end-to-end visuomotor policy learning method leveraging omnidirectional 3D perception with time-aware encoding; and (iv) real-world deployment enabling large-workspace manipulation and omnidirectional obstacle avoidance.}
\label{fig:system}
\end{figure*}

\subsection{System Overview}
% In this section, we present our lightweight teleoperation and learning system. The framework integrates a full-sized humanoid robot platform (Unitree G1 with Dex3-1 dexterous hands) with immersive teleoperation and real-time 3D perception. The system is designed for whole-body data acquisition and policy learning under egocentric 3D observations. In contrast to fixed-base or tabletop manipulation setups, our system keeps all joints enabled and operates from an egocentric LiDAR stream.
In this section, we present a framework that enables large-workspace manipulation for humanoid robots through omnidirectional 3D perception. To overcome the viewpoint constraints of conventional visuomotor policies, our approach leverages panoramic LiDAR observations within an imitation learning pipeline. We first develop a whole-body teleoperation system for efficient collection of demonstration data. We then train OmniDP, an end-to-end visuomotor policy that processes historical sequences of panoramic point clouds using a time-aware encoding mechanism. The system overview is provided in Fig.~\ref{fig:system}.

\subsection{OmniDP}

    \textbf{Architecture of OmniDP.} 
    OmniDP comprises a perception encoder and an action decoder. The encoder processes panoramic point clouds using a pyramid convolutional encoder~\cite{2025idp3} enhanced with time-aware attention, producing a compact global feature that captures both spatial structure and temporal context. This feature conditions a Diffusion Policy~\cite{2025dp} decoder, which iteratively denoises random noise into coordinated action sequences. The end-to-end framework directly maps omnidirectional observations to whole-body actions without intermediate state estimation.

    \begin{figure*}[t]
        \centering
        % \vspace{0.3cm}
        \includegraphics[width=\linewidth]{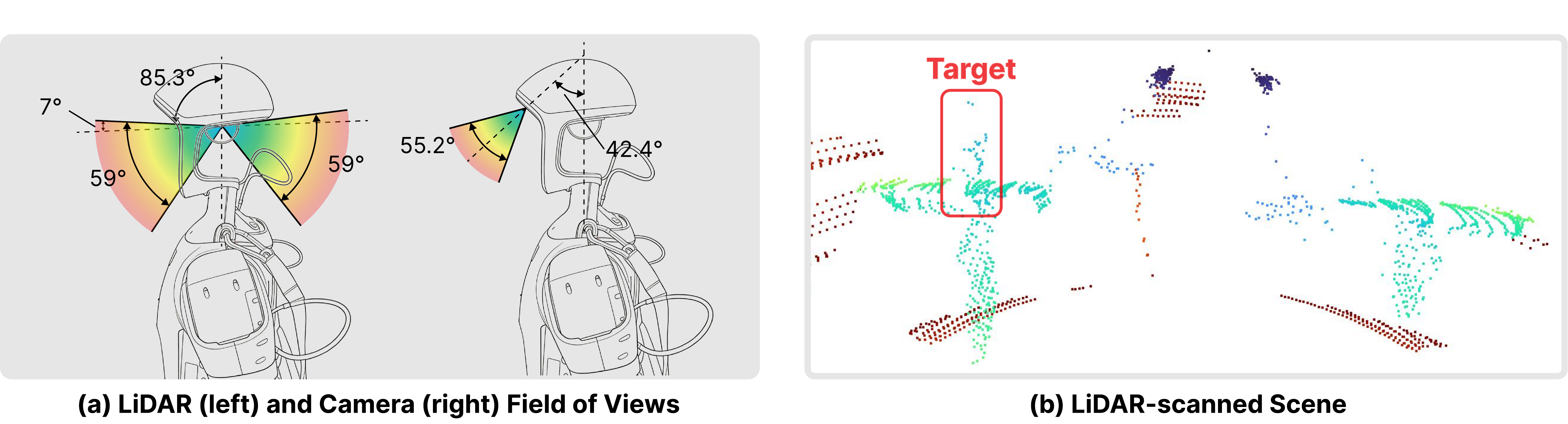}
        \caption{\textbf{Comparison of the field of views of LiDAR and camera.} (a) The panoramic LiDAR provides 360° horizontal coverage, significantly surpassing the narrow field of view of the head-mounted RGB-D camera. (b) This enables the robot to perceive global scene context and locate target objects even when they lie outside the camera's visual blind spots, which is essential for large-workspace manipulation tasks.}
        \label{fig:comparison}
    \end{figure*}

    \textbf{Omnidirectional Perception for Large-Workspace Manipulation.}
    Most existing 3D visuomotor policies operate on third-view~\cite{2024dp3} or egocentric~\cite{2025idp3} point clouds from depth cameras, which suffer from a narrow field of view. This restricts the effective manipulation space to the camera's immediate vicinity, forcing frequent repositioning when objects are distributed across a wider area. As illustrated in Fig.~\ref{fig:comparison}, which compares the visual coverage of LiDAR and RGB-D cameras, we instead adopt panoramic LiDAR to capture complete 360° point clouds at each timestep. This provides full spatial awareness of the robot's surroundings, enabling it to reach and manipulate objects anywhere around its body without moving. It also allows collision-free bimanual coordination in cluttered environments by maintaining continuous awareness of surrounding obstacles throughout task execution.
    
    \textbf{Time-Aware Attention Pooling (TAP).}
    Point clouds generated by LiDAR sensors are often sparse and temporally flickering, making standard global max pooling over single-frame observations unstable. To enhance robustness, we introduce a time-aware attention pooling mechanism for point cloud encoding. Specifically, we aggregate point clouds over a short temporal window and augment each point with a normalized relative timestamp. Per-point spatial features are extracted using a point cloud encoder, while a small MLP predicts attention weights conditioned on the temporal offsets, encouraging the model to emphasize more recent observations. The final global feature is computed as a weighted sum over all points, yielding a temporally smoothed representation that improves stability for policy learning.
    
   \textbf{Point Cloud Processing.}
   Raw panoramic LiDAR point clouds exhibit non-uniform density, with points densely distributed near the sensor and sparse at a distance. To ensure sufficient point density for nearby objects within the robot's manipulation range while maintaining real-time inference efficiency, we design a lightweight preprocessing pipeline. Specifically, we crop out points beyond 1.3 meters from the Lidar sensor based on the robot’s arm reachability prior, where the sparse sampling provides limited utility for manipulation tasks. The remaining points are then uniformly downsampled to 4096 points per frame, balancing geometric fidelity with computational efficiency for 3D encoding.

   \textbf{Learning and Deployment.}
   We train OmniDP using human demonstration data collected via the whole-body teleoperation system described in Sec.~\ref{sec:tele}. To achieve coordinated whole-body control, the model takes as input the 43-dimensional joint states covering all joints of the humanoid as proprioceptive information, together with panoramic point clouds from the head-mounted LiDAR as visual perception. The supervision target is the 28-dimensional joint angles of the upper body, including arms and dexterous hands. During deployment, the learned policy generates commands only for the upper-body joints, while the lower body and waist are controlled by a pre-trained HOMIE~\cite{2025homie} algorithm for stable locomotion. Notably, our point clouds are represented in the LiDAR's egocentric frame, which eliminates the need for extrinsic calibration during deployment and enhances the adaptability of the policy across different environments.

\subsection{Whole-body Teleoperation System}\label{sec:tele}
    \textbf{Hardware Configuration.}
    We deploy a Unitree G1 humanoid robot featuring 29-DoF body and two 7-DoF Dex3-1 dexterous hands, resulting in 43 actively controlled joints. Unlike upper-body-only configurations, all joints including legs, waist, arms, and hands remain enabled during operation. This full-body actuation allows coordinated loco-manipulation behaviors and natural motion execution during teleoperation.
    
    \textbf{Panoramic Lidar Perception.}
    Our perception system centers on a head-mounted Livox MID-360 LiDAR, rigidly installed to align with the operator’s egocentric viewpoint and provide panoramic 3D observations.
    The LiDAR provides a $360^\circ$ horizontal and $-7^\circ$ to $52^\circ$ vertical field-of-view (FoV), generating up to 200k points per second at 10~Hz.
    The wide FoV and non-repetitive scanning pattern improve spatial coverage within short temporal windows, which enhances geometric consistency in cluttered indoor environments. Instead of relying on multi-camera fusion or structured-light depth sensors, we directly utilize raw LiDAR point clouds as the primary visual input modality.
    In simulation, the technology stack of LiDAR is based on DISCOVERSE\cite{2025discoverse}, enabling realistic panoramic point cloud generation.
    
    \textbf{Lightweight Whole-body Teleoperation Device.}
    To enable large-scale data acquisition, we build upon the HOMIE \cite{2025homie} loco-manipulation framework, adapting it to support immersive XR-based teleoperation on the Unitree G1.
    Rather than using isomorphic exoskeleton hardware, our system requires only a standard head-mounted display (HMD) and handheld XR (VR and AR) controllers (Meta Quest 3). This significantly simplifies deployment while retaining full-body controllability.
    The operator’s head motion defines the reference frame for perception, while controller poses specify target end-effector poses. Locomotion and torso adjustments are commanded via joystick inputs, enabling simultaneous manipulation and base motion.

    % \begin{figure}[t]
    %     \centering
    %     \includegraphics[width=\linewidth]{Images/real_task.pdf}
    %     \caption{\textbf{Real-world experiments and snapshots.}}
    %     \label{fig:real}
    % \end{figure}

    \textbf{Task-Space Upper Limb Control.}
    Given target wrist poses from the XR interface, upper limb motion is generated through constrained inverse kinematics (IK).
    Each arm is modeled as a 7-DoF serial chain derived from the Unitree G1 URDF file. Instead of employing classical Jacobian pseudo-inverse methods, we formulate IK as a constrained nonlinear optimization problem. The objective balances:
    1) task-space position and orientation tracking,
    2) joint regularization,  
    and 3) temporal smoothness.
    Joint position limits are explicitly imposed as bound constraints.

    The optimization is solved using interior point optimizer, yielding stable and physically consistent joint configurations even under redundant and multi-objective settings. The optimized configuration $q^*$ is forwarded to the low-level joint controller for execution.

        \begin{table*}[t]
  \vspace{5pt}
  \centering
  \caption{\textbf{Performance comparison.} Our method consistently outperforms existing RGB-based and depth-to-point-cloud baselines across all simulated and real-world tasks, with particularly significant advantages in scenarios where target objects lie outside the camera's field of view.}
  \label{tab: main}
  \resizebox{0.98\textwidth}{!}{ 
      \begin{tabular}{lccccccc}
        \toprule
        & \multicolumn{2}{c}{Simulation} & \multicolumn{4}{c}{Real World} &   \\
        \cmidrule(lr){2-3} \cmidrule(lr){4-7}
        Method & Pick \& Place & Pour (OV) & Pick \& Place & Hand Over (OV) & Pour (OV) & Wipe (OV) & Total \\
        \midrule
        DP & 10/20 & 0/20 & 8/20 & 0/20 & 0/20 & 0/20 & 18/120 \\
        DP3 & 13/20 & 0/20 & 9/20 & 0/20 & 0/20 & 0/20 & 22/120 \\
        iDP3 & 14/20 & 0/20 & 11/20 & 0/20 & 0/20 & 0/20 & 25/120 \\
        \midrule
        \textbf{Ours} & \textbf{16/20} & \textbf{12/20} & \textbf{15/20} & \textbf{12/20} & \textbf{11/20} & \textbf{16/20} & \textbf{82/120} \\
        \bottomrule
      \end{tabular}
    }
\end{table*}

    \textbf{Motion Mapping.}
    Controller poses relative to the HMD frame are mapped to robot end-effector targets through rigid-body transformations in $SE(3)$:
    \begin{equation}
    \boldsymbol{T} =
    \begin{bmatrix}
    \boldsymbol{R} & \boldsymbol{t} \\
    \boldsymbol{0}^\top & 1
    \end{bmatrix},
    \end{equation}
    where $\boldsymbol{R} \in SO(3)$ and $\boldsymbol{t} \in \mathbb{R}^3$ represent rotation and translation, respectively.
    
    Locomotion commands are defined as
    \begin{equation}
    \boldsymbol{u}_k =
    \begin{bmatrix}
    v_{x,k},\ v_{y,k},\ \Delta \theta_k,\ h_k
    \end{bmatrix}^\top,
    \end{equation}
    where $v_{x,k}$ and $v_{y,k}$ denote planar velocities, $\Delta \theta_k$ denotes the yaw increment, and $h_k$ denotes the torso height adjustment. This formulation enables coordinated base motion during manipulation.

     \textbf{Data Collection.} During teleoperation, we record synchronized observation–action trajectories. Observations consist of real-time processed LiDAR point clouds (distance-based cropping and uniform downsampling) and the 43-DoF proprioceptive joint positions. 
    We utilize the TeleVuer open-source framework to establish real-time communication between the XR device and the controlled robot, supporting interactive teleoperation under high-throughput visual data streams, such as high-density 3D point clouds.
    Unlike image-centric pipelines, our dataset preserves full 3D spatial structure, enabling downstream policy learning directly in geometric space.

    \begin{figure}[t]
        \centering
        \includegraphics[width=\linewidth]{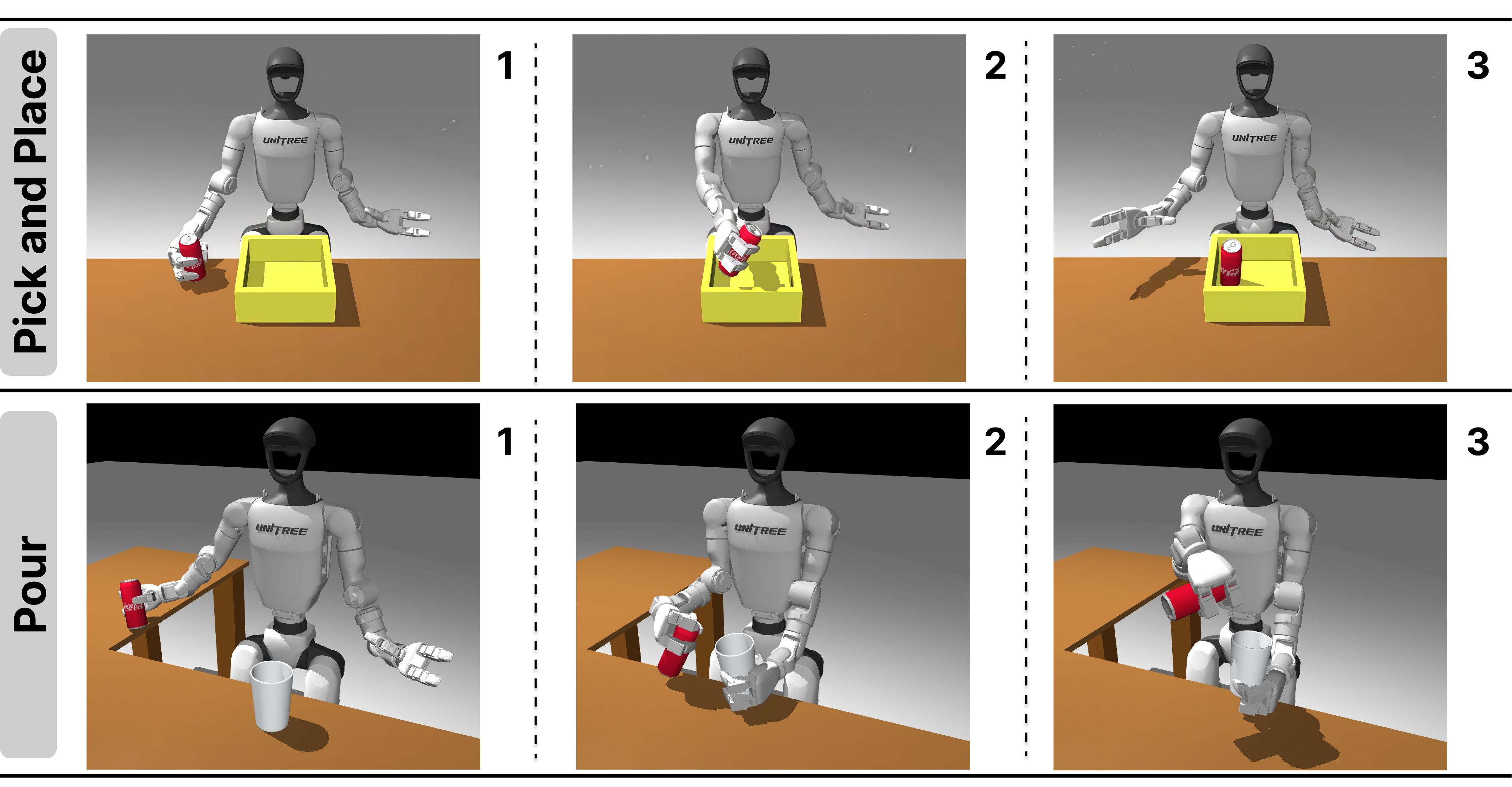}
        \caption{\textbf{Simulation experiments and snapshots.} We evaluate OmniDP on two simulated tasks, demonstrating its effectiveness in large-workspace manipulation scenarios with panoramic perception.}
        \label{fig:sim}
    \end{figure}

\section{Experiments}

\begin{figure}[t]
        \centering
        \includegraphics[width=0.48\textwidth]{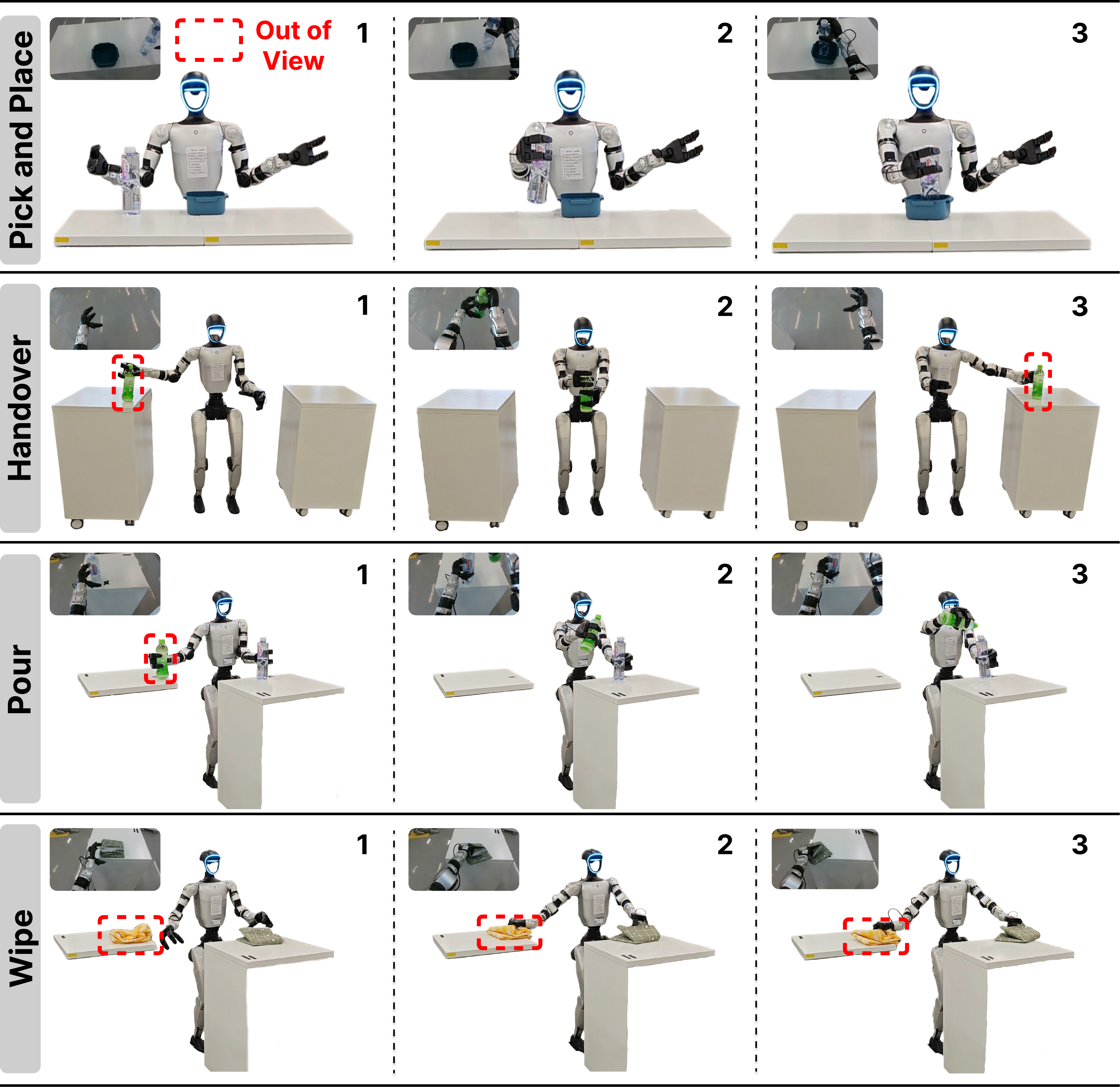} 
        \caption{\textbf{Real-world experiments and snapshots.} We further validate OmniDP on four real-world tasks, confirming its robust performance in real-world environments and out-of-view manipulation scenarios.}
        \label{fig:real}
    \end{figure}

\textbf{Tasks.}
In simulation, we design two simulation tasks using the MuJoCo ~\cite{2012mujoco} physics engine: (1) Pick \& Place, and (2) Pour (OV), as shown in Fig.~\ref{fig:sim}. These tasks are constructed to require large-scale spatial awareness, where critical objects and interactions frequently occur outside the field of view of a fixed camera, leading to partial observability.
To further validate our approach in real-world settings, we additionally introduce two tasks, (3) Hand Over (OV), and (4) Wipe (OV), as shown in Fig.~\ref{fig:real}. Tasks marked with ``OV" (Out of View) require panoramic spatial awareness, where key objects may lie outside the field of view of a fixed camera.

\textbf{Baselines.}
We evaluate our approach against three diffusion-based manipulation policies: DP ~\cite{2025dp}, DP3 ~\cite{2024dp3}, and iDP3 ~\cite{2025idp3}. These methods share a common diffusion-policy backbone but differ in their visual observation modalities and encoder architectures.
By benchmarking under a unified diffusion framework, we aim to systematically examine how variations in visual input representation and feature encoding influence policy performance, particularly in scenarios requiring panoramic or large-scale spatial perception. Task success rate is adopted as the primary evaluation metric.
It is important to note that the comparison with DP / DP3 / iDP3 is not intended to demonstrate the superiority of our algorithm, but rather to validate the impact of the `FOV limitation'.

\textbf{Robot Setup.}
All experiments are conducted on a Unitree G1 humanoid platform equipped with Unitree Dex3-1 power-controlled dexterous hands. A Livox MID-360 LiDAR and an Intel RealSense D435i camera are rigidly mounted on the robot head for exteroceptive perception.
Point clouds are obtained either directly from the LiDAR or converted from depth images via camera intrinsics. During policy deployment, all 29 DoFs of the robot are fully actuated, comprising 14 DoFs in the arms, 3 DoFs in the waist, and 12 DoFs in the legs.

\textbf{Evaluation Metrics.}
We evaluate all methods primarily by task success rate. For each task, we run each model 20 times under consistent initial conditions and report the number of successful trials.  For collision-free manipulation tasks conducted in cluttered environments, we additionally report the collision rate, defined as the proportion of trials where the robot makes physical contact with surrounding obstacles during execution. This metric captures the robot's ability to leverage environmental awareness for safe whole-body coordination, complementing the success rate which focuses on task completion.

    \subsection{Effectiveness of OmniDP}
    We evaluate the effectiveness of OmniDP by comparing it with three diffusion-based visuomotor baselines: Diffusion Policy (DP)~\cite{2025dp} operating on 240 × 424 RGB images, DP3~\cite{2024dp3} and iDP3~\cite{2025idp3}, both using point clouds derived from depth cameras (camera-frame coordinates, downsampled to 4096 points). All methods are tested on two simulated tasks and four real-world tasks, with success rates reported in Table~\ref{tab: main}. For tasks where target objects are located outside the camera's field of view (denoted OV), all baselines largely fail due to their limited perceptual coverage.  In contrast, OmniDP leverages its omnidirectional 360° perception to achieve robust large-workspace manipulation, reaching objects anywhere in the surroundings while maintaining full environmental awareness. For in-view tasks such as Pick \& Place, OmniDP also outperforms the baselines, benefiting from panoramic perception that provides comprehensive scene context for better obstacle avoidance and spatial reasoning, while the time-aware encoding mechanism enhances temporal stability by mitigating transient sensor noise.

    \begin{table}[t]
      \centering
      \vspace{0.5cm}
      \caption{\textbf{Collision-free manipulation evaluation.} Results demonstrate that OmniDP achieves robust omnidirectional obstacle avoidance in cluttered environments.}
      \label{tab: collision}
      \resizebox{0.4\textwidth}{!}{ 
          \begin{tabular}{l|cc}
            \toprule
            Method & Success Rate $\uparrow$ & Collision Rate $\downarrow$ \\
            \midrule
            DP & 0/20 & 20/20 \\
            DP3  & 0/20 & 18/20 \\
            iDP3 & 0/20 & 18/20 \\
            \midrule
            \textbf{Ours} & \textbf{14/20} & \textbf{5/20} \\
            \bottomrule
          \end{tabular}
        }
    \end{table}

    \begin{table}[t]
      \centering
      \caption{\textbf{Generalization capability evaluation.} Results demonstrate that OmniDP maintains consistently high success rates across varying object instances, lighting conditions, and cluttered scenes.}
      \label{tab: generalization}
      \resizebox{0.38\textwidth}{!}{ 
          \begin{tabular}{l|ccc}
            \toprule
            Method & Instance & Lighting & Scene\\
            \midrule
            DP & 3/20 & 4/20 & 2/20\\
            DP3  & 7/20 & 8/20 & 7/20\\
            iDP3 & 12/20 & 12/20 & 10/20\\
            \midrule
            \textbf{Ours} & \textbf{13/20} & \textbf{15/20} & \textbf{12/20}\\
            \bottomrule
          \end{tabular}
        }
    \end{table}
    
    \subsection{Collision-Free Manipulation Evaluation}
\begin{figure}[t]
        \centering
        \vspace{0.5cm}
        \includegraphics[width=\linewidth]{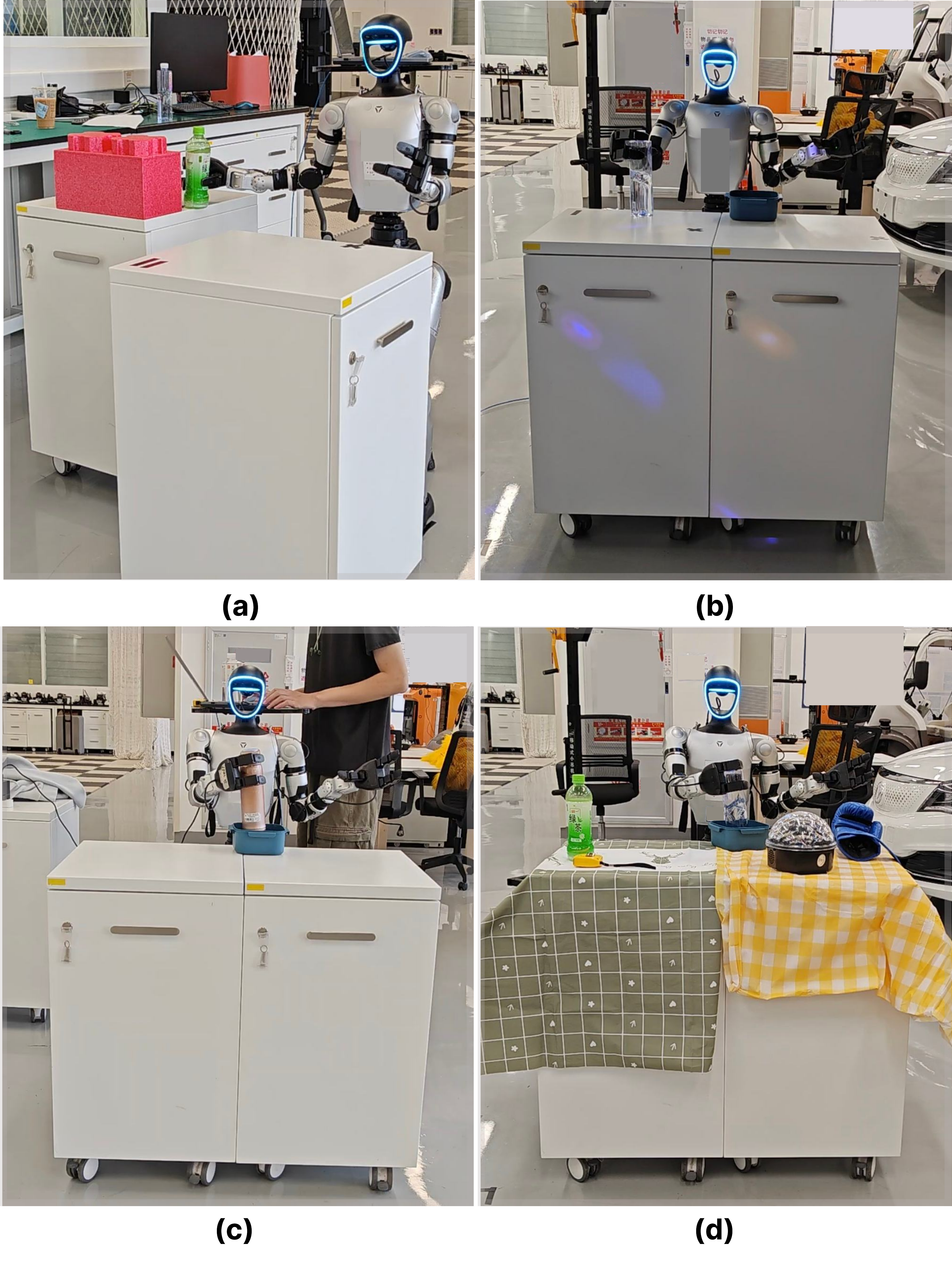}
        \caption{\textbf{Capabilities evaluation.} We test OmniDP under diverse challenging conditions: (a) omnidirectional obstacle avoidance, where obstacles are placed outside the camera's field of view; (b) varying object instances with different shapes and sizes; (c) changing lighting conditions, including low-light and complex lighting; and (d) cluttered environments with multiple distractor objects.}
        \label{fig:capability}
    \end{figure}

    We further evaluate the obstacle avoidance capability of each method during task execution. In this experiment, we place a physical obstacle along the robot's manipulation path, positioned outside the field of view of the head-mounted RGB-D camera, as illustrated in Fig.~\ref{fig:capability}(a). This setup challenges the policy to perceive and avoid obstacles that are not visible to standard egocentric vision but are captured by panoramic LiDAR. We compare OmniDP against the three baselines in terms of success rate and collision rate, with results reported in Table~\ref{tab: collision}. OmniDP demonstrates robust performance and effective omnidirectional obstacle avoidance, successfully completing tasks while avoiding collisions in most trials. In contrast, all baselines largely fail to complete the task, frequently colliding with the obstacle due to their inability to perceive objects outside the limited camera field of view. These results highlight the importance of panoramic perception for safe whole-body manipulation in cluttered environments.

    \subsection{Generalization Capability Evaluation}
    We also assess the generalization ability of OmniDP under varying environmental conditions. Specifically, we test on the Pick \& Place task, the setting where all baselines achieve reasonable performance under nominal conditions, and introduce three types of variations: (1) different object instances with diverse shapes and sizes, (2) changing lighting conditions, and (3) cluttered scenes with distractor objects placed around the workspace, as depicted in Fig.~\ref{fig:capability}(b)-(d). Success rates for each variant are reported in Table~\ref{tab: generalization}. While baseline methods experience noticeable performance drops due to their reliance on narrow-view RGB-D inputs and sensitivity to visual perturbations, OmniDP maintains consistently high success rates across all variations. This robustness stems from its panoramic scene understanding, which provides comprehensive spatial context and is inherently less affected by lighting changes or local clutter, enabling reliable operation in diverse real-world scenarios.

    \subsection{Ablation Study}
    We conduct ablation studies to validate the key components of OmniDP: omnidirectional perception and the time-aware attention pooling (TAP) mechanism. Experiments are performed on the Hand Over task, which demands large-workspace coordination and spatial awareness. As reported in Table~\ref{tab: abla}, removing the panoramic LiDAR input (i.e., relying solely on egocentric depth-based point clouds) leads to complete task failure, highlighting the critical role of 360° scene understanding for such tasks. Furthermore, disabling TAP results in a noticeable decrease in success rate, indicating that temporal smoothing via attention over historical point clouds contributes to more stable policy execution. These results confirm that both panoramic perception and temporal encoding are essential for reliable large-workspace manipulation.

    \begin{table}[htbp]
      \centering
      \caption{\textbf{Ablation study.} The results confirm the necessity of the designed components.}
      \label{tab: abla}
      \resizebox{0.32\textwidth}{!}{ 
          \begin{tabular}{l|c}
            \toprule
            Ablation & Success Rate \\
            \midrule
            OmniDP & 12/20 \\
            w/o Omni. Obs.   & 0/20  \\
            w/o TAP   & 9/20  \\
            \bottomrule
          \end{tabular}
        }
    \end{table}

\section{Conclusion}

In this paper, we present OmniDP, an end-to-end LiDAR-driven visuomotor policy for large-workspace humanoid dexterous manipulation on humanoid robots. By leveraging panoramic LiDAR observations with a time-aware attention pooling mechanism, our approach overcomes the fundamental perceptual limitations of conventional RGB-D based policies, enabling 360° scene awareness and collision-free manipulation in cluttered environments. We also introduce a whole-body teleoperation system for efficient collection of demonstration data involving complex full-body motions. Extensive experiments in both simulation and real-world settings demonstrate that OmniDP achieves robust performance across large-workspace tasks and out-of-view manipulation scenarios, significantly outperforming vision-based baselines. Ablation studies further validate the importance of panoramic perception and temporal encoding for reliable deployment. We believe that integrating omnidirectional 3D perception into visuomotor policy learning offers a promising direction toward humanoid robots capable of operating safely and dexterously in unstructured real-world environments.

\textbf{Future Work.} Future work will extend the system toward whole-body autonomy, leveraging LiDAR to perceive targets beyond the camera’s field of view and using locomotion for active viewpoint acquisition. We also plan to incorporate multi-modal sensing by integrating LiDAR geometry, RGB semantics, and tactile feedback. On the control side, augmenting diffusion policy with reinforcement learning represents a promising direction to improve action refinement and adaptability. Strengthening the policy’s generalization and zero-shot performance on unseen objects, layouts, and tasks will be another key step toward reliable deployment in real-world environments.

% \textbf{Future work.} Future work will extend the system toward whole-body autonomy by using LiDAR to perceive targets beyond the camera’s field of view and employing robot locomotion to actively acquire more informative viewpoints. We also plan to integrate multi-modal feedback by combining LiDAR geometry, RGB semantics and tactile sensing.
% On the control side, we aim to augment diffusion policy with reinforcement learning to improve action refinement and adaptability. A further direction is to enhance generalization and zero-shot performance in novel objects, arrangements and tasks, moving toward a mobile and robust humanoid manipulation system suitable for real-world deployment.

% 可选部分（根据需要取消注释）
% \input{Sections/7_Acknowledge}
% \input{Sections/8_Appendix}

% 参考文献

\balance

\bibliographystyle{ieeetr}
\bibliography{root}

@article{2025visionInAction,
      title={Vision in action: Learning active perception from human demonstrations},
      author={Xiong, Haoyu and Xu, Xiaomeng and Wu, Jimmy and Hou, Yifan and Bohg, Jeannette and Song, Shuran},
      journal={arXiv preprint arXiv:2506.15666},
      year={2025}
    }

@article{2025egomi,
      title={Egomi: Learning active vision and whole-body manipulation from egocentric human demonstrations},
      author={Yu, Justin and Shentu, Yide and Wu, Di and Abbeel, Pieter and Goldberg, Ken and Wu, Philipp},
      journal={arXiv preprint arXiv:2511.00153},
      year={2025}
    }

@article{2024aloha,
      title={Mobile {A}loha: Learning bimanual mobile manipulation with low-cost whole-body teleoperation},
      author={Fu, Zipeng and Zhao, Tony Z and Finn, Chelsea},
      journal={arXiv preprint arXiv:2401.02117},
      year={2024}
    }

@article{2024dreamitate,
      title={Dreamitate: Real-world visuomotor policy learning via video generation},
      author={Liang, Junbang and Liu, Ruoshi and Ozguroglu, Ege and Sudhakar, Sruthi and Dave, Achal and Tokmakov, Pavel and Song, Shuran and Vondrick, Carl},
      journal={arXiv preprint arXiv:2406.16862},
      year={2024}
    }

@article{2025visualImitation,
      title={Visual imitation enables contextual humanoid control},
      author={Allshire, Arthur and Choi, Hongsuk and Zhang, Junyi and McAllister, David and Zhang, Anthony and Kim, Chung Min and Darrell, Trevor and Abbeel, Pieter and Malik, Jitendra and Kanazawa, Angjoo},
      journal={arXiv preprint arXiv:2505.03729},
      year={2025}
    }

@article{2024octo,
      title={Octo: An open-source generalist robot policy},
      author={Team, Octo Model and Ghosh, Dibya and Walke, Homer and Pertsch, Karl and Black, Kevin and Mees, Oier and Dasari, Sudeep and Hejna, Joey and Kreiman, Tobias and Xu, Charles and others},
      journal={arXiv preprint arXiv:2405.12213},
      year={2024}
    }

@article{2024humanoidAssistance,
      title={Humanoid-human sit-to-stand-to-sit assistance},
      author={Lef{\`e}vre, Hugo and Chaki, Tomohiro and Kawakami, Tomohiro and Tanguy, Arnaud and Yoshiike, Takahide and Kheddar, Abderrahmane},
      journal={IEEE Robotics and Automation Letters},
      volume={10},
      number={2},
      pages={1521--1528},
      year={2024}
    }

@article{2025behaviorRobotSuite,
      title={Behavior robot suite: Streamlining real-world whole-body manipulation for everyday household activities},
      author={Jiang, Yunfan and Zhang, Ruohan and Wong, Josiah and Wang, Chen and Ze, Yanjie and Yin, Hang and Gokmen, Cem and Song, Shuran and Wu, Jiajun and Fei-Fei, Li},
      journal={arXiv preprint arXiv:2503.05652},
      year={2025}
    }

@article{2025cola,
      title={Learning Human-Humanoid Coordination for Collaborative Object Carrying},
      author={Du, Yushi and Li, Yixuan and Jia, Baoxiong and Lin, Yutang and Zhou, Pei and Liang, Wei and Yang, Yanchao and Huang, Siyuan},
      journal={arXiv preprint arXiv:2510.14293},
      year={2025}
    }

@article{2025robopanoptes,
      title={Robopanoptes: The all-seeing robot with whole-body dexterity},
      author={Xu, Xiaomeng and Bauer, Dominik and Song, Shuran},
      journal={arXiv preprint arXiv:2501.05420},
      year={2025}
    }

@inproceedings{2023robotStructure,
      title={Robot structure prior guided temporal attention for camera-to-robot pose estimation from image sequence},
      author={Tian, Yang and Zhang, Jiyao and Yin, Zekai and Dong, Hao},
      booktitle={Proceedings of the IEEE/CVF Conference on Computer Vision and Pattern Recognition},
      pages={8917--8926},
      year={2023}
    }

@article{tian2024viewinvariant,
      title={View-invariant policy learning via zero-shot novel view synthesis},
      author={Tian, Stephen and Wulfe, Blake and Sargent, Kyle and Liu, Katherine and Zakharov, Sergey and Guizilini, Vitor and Wu, Jiajun},
      journal={arXiv preprint arXiv:2409.03685},
      year={2024}
    }

@article{2024robotSeeRobotDo,
      title={Robot see robot do: Imitating articulated object manipulation with monocular 4{D} reconstruction},
      author={Kerr, Justin and Kim, Chung Min and Wu, Mingxuan and Yi, Brent and Wang, Qianqian and Goldberg, Ken and Kanazawa, Angjoo},
      journal={arXiv preprint arXiv:2409.18121},
      year={2024}
    }

@inproceedings{2024pcdImprove,
      title={Point cloud models improve visual robustness in robotic learners},
      author={Peri, Skand and Lee, Iain and Kim, Chanho and Fuxin, Li and Hermans, Tucker and Lee, Stefan},
      booktitle={2024 IEEE International Conference on Robotics and Automation (ICRA)},
      pages={17529--17536},
      year={2024}
    }

@article{2023visualPolicy,
      title={Visual-policy learning through multi-camera view to single-camera view knowledge distillation for robot manipulation tasks},
      author={Acar, Cihan and Binici, Kuluhan and Tekirda{\u{g}}, Alp and Wu, Yan},
      journal={IEEE Robotics and Automation Letters},
      volume={9},
      number={1},
      pages={691--698},
      year={2023}
    }

@inproceedings{2025PLKCalib,
      title={P{LK}-{C}alib: Single-shot and Target-less {LiDAR}-Camera Extrinsic Calibration using Pl{\"u}cker Lines},
      author={Zhang, Yanyu and Xu, Jie and Ren, Wei},
      booktitle={2025 IEEE/RSJ International Conference on Intelligent Robots and Systems (IROS)},
      pages={16091--16097},
      year={2025}
    }

@inproceedings{2025dur360bev,
      title={Dur360{BEV}: A Real-World 360-Degree Single Camera Dataset and Benchmark for Bird-Eye View Mapping in Autonomous Driving},
      author={Wenke, E and Yuan, Chao and Li, Li and Sun, Yixin and Gaus, Yona Falinie A and Atapour-Abarghouei, Amir and Breckon, Toby P},
      booktitle={2025 IEEE International Conference on Robotics and Automation (ICRA)},
      pages={3737--3744},
      year={2025}
    }

@inproceedings{2024fisheyebevseg,
      title={Fisheyebevseg: Surround view fisheye cameras based bird's-eye view segmentation for autonomous driving},
      author={Yogamani, Senthil and Unger, David and Narayanan, Venkatraman and Kumar, Varun Ravi},
      booktitle={Proceedings of the IEEE/CVF Conference on Computer Vision and Pattern Recognition},
      pages={1331--1334},
      year={2024}
    }

@inproceedings{2024omnidirectional,
      title={Omnidirectional dense slam for back-to-back fisheye cameras},
      author={Xie, Weijian and Chu, Guanyi and Qian, Quanhao and Yu, Yihao and Zhai, Shangjin and Chen, Danpeng and Wang, Nan and Bao, Hujun and Zhangv, Guofeng},
      booktitle={2024 IEEE International Conference on Robotics and Automation (ICRA)},
      pages={1653--1660},
      year={2024}
    }

@article{2025yuto,
      title={{YUTO MMS}: A comprehensive SLAM dataset for urban mobile mapping with tilted LiDAR and panoramic camera integration},
      author={Zhang, Yiujia and Ahmadi, SeyedMostafa and Kang, Jungwon and Arjmandi, Zahra and Sohn, Gunho},
      journal={The International Journal of Robotics Research},
      volume={44},
      number={1},
      pages={3--21},
      year={2025},
      publisher={SAGE Publications Sage UK: London, England}
    }

@article{2024learning,
      title={Learning robust autonomous navigation and locomotion for wheeled-legged robots},
      author={Lee, Joonho and Bjelonic, Marko and Reske, Alexander and Wellhausen, Lorenz and Miki, Takahiro and Hutter, Marco},
      journal={Science Robotics},
      volume={9},
      number={89},
      pages={eadi9641},
      year={2024},
      publisher={American Association for the Advancement of Science}
    }

@article{2024navila,
      title={Navila: Legged robot vision-language-action model for navigation},
      author={Cheng, An-Chieh and Ji, Yandong and Yang, Zhaojing and Gongye, Zaitian and Zou, Xueyan and Kautz, Jan and B{\i}y{\i}k, Erdem and Yin, Hongxu and Liu, Sifei and Wang, Xiaolong},
      journal={arXiv preprint arXiv:2412.04453},
      year={2024}
    }

@article{2025safeLocomotion,
      title={End-to-End Humanoid Robot Safe and Comfortable Locomotion Policy},
      author={Wang, Zifan and Yang, Xun and Zhao, Jianzhuang and Zhou, Jiaming and Ma, Teli and Gao, Ziyao and Ajoudani, Arash and Liang, Junwei},
      journal={arXiv preprint arXiv:2508.07611},
      year={2025}
    }

@inproceedings{chuang2025active,
  title={Active vision might be all you need: Exploring active vision in bimanual robotic manipulation},
  author={Chuang, Ian and Lee, Andrew and Gao, Dechen and Naddaf-Sh, M-Mahdi and Soltani, Iman},
  booktitle={2025 IEEE International Conference on Robotics and Automation (ICRA)},
  pages={7952--7959},
  year={2025}
}

@article{2025homie,
  title={{HOMIE}: Humanoid loco-manipulation with isomorphic exoskeleton cockpit},
  author={Ben, Qingwei and Jia, Feiyu and Zeng, Jia and Dong, Junting and Lin, Dahua and Pang, Jiangmiao},
  journal={arXiv preprint arXiv:2502.13013},
  year={2025}
}

@article{2025RLmani,
      title={Sim-to-real reinforcement learning for vision-based dexterous manipulation on humanoids},
      author={Lin, Toru and Sachdev, Kartik and Fan, Linxi and Malik, Jitendra and Zhu, Yuke},
      journal={arXiv preprint arXiv:2502.20396},
      year={2025}
    }

@article{2025cotraining,
      title={Sim-and-real co-training: A simple recipe for vision-based robotic manipulation},
      author={Maddukuri, Abhiram and Jiang, Zhenyu and Chen, Lawrence Yunliang and Nasiriany, Soroush and Xie, Yuqi and Fang, Yu and Huang, Wenqi and Wang, Zu and Xu, Zhenjia and Chernyadev, Nikita and others},
      journal={arXiv preprint arXiv:2503.24361},
      year={2025}
    }

@inproceedings{2025monocular,
      title={Monocular One-Shot Metric-Depth Alignment for RGB-Based Robot Grasping},
      author={Guo, Teng and Huang, Baichuan and Yu, Jingjin},
      booktitle={2025 IEEE/RSJ International Conference on Intelligent Robots and Systems (IROS)},
      pages={642--649},
      year={2025}
    }

@article{2025dp,
      title={Diffusion {P}olicy: Visuomotor policy learning via action diffusion},
      author={Chi, Cheng and Xu, Zhenjia and Feng, Siyuan and Cousineau, Eric and Du, Yilun and Burchfiel, Benjamin and Tedrake, Russ and Song, Shuran},
      journal={The International Journal of Robotics Research},
      volume={44},
      number={10-11},
      pages={1684--1704},
      year={2025},
      publisher={Sage Publications Sage UK: London, England}
    }

@article{2024dp3,
      title={3{D} {D}iffusion {P}olicy: Generalizable visuomotor policy learning via simple 3d representations},
      author={Ze, Yanjie and Zhang, Gu and Zhang, Kangning and Hu, Chenyuan and Wang, Muhan and Xu, Huazhe},
      journal={arXiv preprint arXiv:2403.03954},
      year={2024}
    }

@inproceedings{2025idp3,
      title={Generalizable humanoid manipulation with 3{D} diffusion policies},
      author={Ze, Yanjie and Chen, Zixuan and Wang, Wenhao and Chen, Tianyi and He, Xialin and Yuan, Ying and Peng, Xue Bin and Wu, Jiajun},
      booktitle={2025 IEEE/RSJ International Conference on Intelligent Robots and Systems (IROS)},
      pages={2873--2880},
      year={2025}
    }

@article{2025manivid,
      title={Mani{VID}-3{D}: Generalizable View-Invariant Reinforcement Learning for Robotic Manipulation via Disentangled 3{D} Representations},
      author={Li, Zheng and Qu, Pei and Jia, Yufei and Zhou, Shihui and Ge, Haizhou and Cao, Jiahang and Zhou, Jinni and Zhou, Guyue and Ma, Jun},
      journal={arXiv preprint arXiv:2509.11125},
      year={2025}
    }

@article{2025ISSpolicy,
      title={{ISS} {P}olicy: Scalable Diffusion Policy with Implicit Scene Supervision},
      author={Xia, Wenlong and Zhang, Jinhao and Zhang, Ce and Wang, Yaojia and Gong, Youmin and Mei, Jie},
      journal={arXiv preprint arXiv:2512.15020},
      year={2025}
    }

@article{2026informationFiltering,
      title={Information Filtering via Variational Regularization for Robot Manipulation},
      author={Zhang, Jinhao and Xia, Wenlong and Wang, Yaojia and Zhou, Zhexuan and Li, Huizhe and Lai, Yichen and Song, Haoming and Gong, Youmin and Me, Jie},
      journal={arXiv preprint arXiv:2601.21926},
      year={2026}
    }

@article{2025omniperception,
    title={Omni-{P}erception: Omnidirectional collision avoidance for legged locomotion in dynamic environments},
    author={Wang, Zifan and Ma, Teli and Jia, Yufei and Yang, Xun and Zhou, Jiaming and Ouyang, Wenlong and Zhang, Qiang and Liang, Junwei},
    journal={arXiv preprint arXiv:2505.19214},
    year={2025}
}

@article{2025lvi-q,
  title={{LVI-Q}: Robust LiDAR-Visual-Inertial-Kinematic Odometry for Quadruped Robots Using Tightly-Coupled and Efficient Alternating Optimization},
  author={Marsim, Kevin Christiansen and Oh, Minho and Yu, Byeongho and Lee, Seungjae and Nahrendra, I Made Aswin and Lim, Hyungtae and Myung, Hyun},
  journal={IEEE Robotics and Automation Letters},
  year={2025}
}

@article{2026pilot,
  title={{PILOT}: A Perceptive Integrated Low-level Controller for Loco-manipulation over Unstructured Scenes},
  author={Cui, Xinru and Feng, Linxi and Zhou, Yixuan and Han, Haoqi and Liu, Zhe and Wang, Hesheng},
  journal={arXiv preprint arXiv:2601.17440},
  year={2026}
}

@article{2025beamdojo,
  title={Beam{D}ojo: Learning agile humanoid locomotion on sparse footholds},
  author={Wang, Huayi and Wang, Zirui and Ren, Junli and Ben, Qingwei and Huang, Tao and Zhang, Weinan and Pang, Jiangmiao},
  journal={arXiv preprint arXiv:2502.10363},
  year={2025}
}

@inproceedings{2012mujoco,
  title={Mu{J}o{C}o: A physics engine for model-based control},
  author={Todorov, Emanuel and Erez, Tom and Tassa, Yuval},
  booktitle={2012 IEEE/RSJ International Conference on Intelligent Robots and Systems},
  pages={5026--5033},
  year={2012}
}

@inproceedings{2025discoverse,
  title={{DISCOVERSE}: Efficient robot simulation in complex high-fidelity environments},
  author={Jia, Yufei and Wang, Guangyu and Dong, Yuhang and Wu, Junzhe and Zeng, Yupei and Lin, Haonan and Wang, Zifan and Ge, Haizhou and Gu, Weibin and Ding, Kairui and others},
  booktitle={2025 IEEE/RSJ International Conference on Intelligent Robots and Systems (IROS)},
  pages={6272--6279},
  year={2025}
}

\end{document}